\documentclass[letterpaper]{article}
\usepackage{aaai23} 
\usepackage{times} 
\usepackage{helvet} 
\usepackage{courier} 
\usepackage[hyphens]{url} 
\usepackage{graphicx} 
\usepackage{graphicx} 
\usepackage{natbib} 
\usepackage{caption} 
\usepackage{newfloat}
\usepackage{listings}
\usepackage{here}
\DeclareCaptionStyle{ruled}{labelfont=normalfont,labelsep=colon,strut=off} 
\title{Formally Verified Solution Methods for Markov Decision Processes}
\author{
    Maximilian Schäffeler\textsuperscript{\rm 1},
    Mohammad Abdulaziz\textsuperscript{\rm 1,2}
}
\affiliations{
    \textsuperscript{\rm 1}Technische Universität München, Germany\\
    \textsuperscript{\rm 2}King's College London, United Kingdom
}

\usepackage{amsmath, amssymb, amsthm}
\usepackage{mathtools}
\usepackage[ruled,noline,noalgohanging]{algorithm2e}
\usepackage{tabularx}
\usepackage{multirow}
\usepackage{todonotes}
\usepackage{subcaption}

\usepackage[most]{tcolorbox}
\tcbuselibrary{breakable,hooks,skins,theorems}

\let\showlistings\undefined 

\ifdefined\showlistings
\newcommand{\listingref}[1]{#1}
\else
\newcommand{\listingref}[1]{}
\fi

\def\showappendix{} 

\ifdefined\showappendix
\newcommand{\appendixenv}[1]{#1}
\else
\newcommand{\appendixenv}[1]{}
\fi
\renewcommand{\todo}[1]{}

\newtheorem{theorem}{Theorem}
\newtheorem{definition}{Definition}

\DeclareMathOperator*{\argmax}{arg\,max}

\def\returnop{\mathrm{return}}
\def\cP{P}
\def\PP{\mathcal{P}}
\def\cL{L}
\def\GS{G}

\lstdefinelanguage{isabelle}{
    morekeywords={record,type_synonym,definition,fun,function,primrec,where,lemma,theorem,unfolding,by,shows,assumes,and,datatype,using,abbreviation
,moreover,have,hence,thus,qed,proof,let,ultimately,show,next,in}
    , sensitive=true
    , showstringspaces=false
    , framerule=0pt
    , xleftmargin=2em
    , numbers=left
    , numberstyle=\ttfamily\tiny
    , firstnumber=1
    , stepnumber=2
    , basicstyle=\ttfamily\tiny
    , breaklines=true
    , showspaces=false
    , morecomment=[l]{--}
    , morecomment=[s]{(*}{*)}
    , commentstyle=\color{gray}
    , morestring=[b]"
    , literate={\\<times>}{{$\times$}}{1} {\\<equiv>}{{$\equiv$}}{1} {\\<forall>}{{$\forall$}}{1} {\\<exists>}{{$\exists$}}{1} {\\<and>}{{$\land$}}{1}
        {\\<in>}{{$\in$}}{1} {\\<Rightarrow>}{{$\Rightarrow$}}{1} {\\<lambda>}{{$\lambda$}}{1} {::}{{$::$}}{1}
        {\\<subseteq>}{{$\subseteq$}}{1} 
        {\\<circ>}{{$\circ$}}{1} 
        {\\<^sub>m}{{$_m$}}{1} 
        {\\<longleftrightarrow>}{{$\longleftrightarrow$}}{3}
        {\\<pi>}{{$\pi$}}{1} {\\<delta>}{{$\delta$}}{1} 
        {\\<omega>}{{$\omega$}}{1}
        {\\<bind>}{$\bindop$}{1}
        {\\<And>}{$\bigwedge$}{1}
        {\\<dots>}{$\ldots$}{3}
        {\\<nu>}{$\nu$}{1}
        {\\<Sum>}{$\sum$}{2}
        {\\<integral>}{$\int$}{1}
        {\\<partial>}{$\partial$}{2}
        {\\<T>}{{$\mathcal T$}}{1}
        {\\<P>}{{$\mathcal P$}}{1}
        {\\<L>}{{$\mathcal L$}}{1}
        {\\<Squnion>}{{$\bigsqcup$}}{2}
        {\\<lbrakk>}{{$\llbracket$}}{1} {\\<rbrakk>}{{$\rrbracket$}}{1}
        {\\<Longrightarrow>}{{$\Longrightarrow$}}{3}
        {\\<longlonglongrightarrow>}{{$\xrightarrow{\hspace*{0.4cm}}$}}{4}
        {\\<not>}{{$\lnot$}}{1} {\\<le>}{{$\le$}}{1} {\\<rightharpoonup>}{{$\rightharpoonup$}}{2}
        {\\<^sub>\\<V>}{{$_{\mathcal V}$}}{1}
        {\\<^sub>H}{{$_{\texttt H}$}}{1}
        {\\<^sub>R}{{$_{\texttt R}$}}{1}
        {\\<^sub>D}{{$_{\texttt D}$}}{1}
        {\\<^sub>U}{{$_{\texttt U}$}}{1}
        {\\<^sub>b}{{$_{\texttt{b}}$}}{1}
        {\\<lparr>}{{$\llparenthesis$}}{1} {\\<rparr>}{{$\rrparenthesis$}}{1}
        {\\<leftarrow>}{{$\leftarrow$}}{1} {\\<^sub>\\<O>}{{$_{\mathcal O}$}}{1} {\\<^sub>I}{{$_{\texttt{I}}$}}{1}
        {\\<^sub>G}{{$_{\texttt{G}}$}}{2} {\\<phi>}{{$\varphi$}}{1} {\\<Phi>}{{$\Phi$}}{1} {\\<psi>}{{$\psi$}}{1} {\\<Psi>}{{$\Psi$}}{1}
        {\\<^sub>S}{{$_{\texttt S}$}}{1} {\\<inverse>}{{$^{-1}$}}{1} {\\<^sub>O}{{$_{\texttt O}$}}{1} {\\<^bold>\\<And>}{{$\bm\bigwedge$}}{1}
        {\\<^bold>\\<or>}{{$\bm\lor$}}{1} {\\<^sub>G}{{$_{\texttt G}$}}{1} {\\<Pi>}{{$\Pi$}}{1} {\\<^sub>I}{{$_{\texttt I}$}}{1} {\\<noteq>}{{$\neq$}}{1}
        {\\<bottom>}{{$\bot$}}{1}
        {\\<^sub>+}{{$_\texttt +$}}{1}
        {\\<^bold>\\<and>}{{$\bm\land$}}{1} {\\<^bold>\\<not>}{{$\bm\lnot$}}{1}
        {\\<^sub>1}{{$_\texttt 1$}}{1} {\\<^sub>2}{{$_2$}}{1} {\\<A>}{{$\mathcal A$}}{1} {\\<Turnstile>}{{$\models$}}{2} {\\<^sub>\\<forall>}{{$_\forall$}}{1}
        {\\<^sub>0}{{$_0$}}{1} {\\<tau>}{{$\tau$}}{1}  {\\<^sub>\\<Omega>}{{$_\Omega$}}{1} {\\<^sub>V}{{$_V$}}{1} {\\<^bold>\\<Or>}{{$\bm\bigvee$}}{1}
        {\\<^sub>P}{{$_\texttt P$}}{1}
        {\\<^sub>X}{{$_\texttt X$}}{1}
        {\\<^sub>M}{{$_\texttt M$}}{1}
        {\\<^sub>L}{{$_\texttt L$}}{1}
        {\\<longrightarrow>}{{$\longrightarrow$}}{2}
        {\\<or>}{{$\lor$}}{1}
        {\\<^sub>\\<pi>}{{$_\pi$}}{1}
        {\\<^sub>s}{{$_s$}}{1}
        {\\<^sub>t}{{$_t$}}{1}
        {\\<^sub>a}{{$_a$}}{1}
        {\\<^sub>r}{{$_r$}}{1}
        {\\<^sub>t}{{$_t$}}{1}
        {\\<^sub>e}{{$_e$}}{1}
        {\\<^sub>n}{{$_n$}}{1}
        {\\<^sub>d}{{$_d$}}{1}
        {\\<^sub>i}{{$_i$}}{1}
        {\\<^sub>v}{{$_v$}}{1}
        {\\<^sub>j}{{$_j$}}{1}
        {\\<^sub>b}{{$_\texttt b$}}{1}
        {\\<inter>}{{$\cap$}}{1}
        {\\<union>}{{$\cup$}}{1}
        {\\<Union>}{{$\bigcup$}}{1}
        {\\<^sup>c\\<TTurnstile>\\<^sub>=}{{${}^c\models_=$}}{1}
        {\\<open>}{{<}}{1}
        {\\<close>}{{>}}{1}
        {\\<langle>}{{$\langle$}}{1}
        {\\<rangle>}{{$\rangle$}}{1}
        {\\<ge>}{{$\ge$}}{1}
}

\newtcblisting[auto counter]{IsabelleSnippet}[2][]{
    listing options={
        language=isabelle
    }
    , listing only
    , enhanced jigsaw
    , breakable
    , top=-.2em
    , right=0pt
    , bottom=-.2em
    , left=0pt
    , boxrule=0pt
    , toprule=1pt
    , bottomrule=1pt
    , titlerule=1pt
    , colframe=black
    , sharp corners
    , colbacktitle=white
    , coltitle=black
    , fonttitle=\tiny
    , colback=white
    , fontupper=\ttfamily\tiny
    , before upper={\parindent0em}
    , title=Listing \thetcbcounter: #2
    , #1
}

\begin{document}
\maketitle

\begin{abstract}
We formally verify executable algorithms for solving Markov decision processes (MDPs) in the interactive theorem prover Isabelle/HOL.
We build on existing formalizations of probability theory to analyze the expected total reward criterion on
finite and infinite-horizon problems.
Our developments formalize the Bellman equation and give conditions under which optimal policies exist.
Based on this analysis, we verify dynamic programming algorithms to solve tabular MDPs.
We evaluate the formally verified implementations experimentally on standard problems, compare them with state-of-the-art systems, and show that they are practical.
\end{abstract}

\newcommand{\const}[2]{\newcommand{#1}{\textrm{#2}}}
\newcommand{\type}[2]{\newcommand{#1}{\textrm{#2}}}
\newcommand{\key}[2]{\newcommand{#1}{\textbf{#2}}}

\newcommand{\typef}[1]{\textrm{#1}}
\newcommand{\constf}[1]{\textrm{#1}}

\newcommand{\Kzero}{\textsf{P}_\textsf{X}}
\newcommand{\Kstep}{\constf{P}_\textsf{step}}
\newcommand{\Lact}{\constf{L}_\textsf{act}}
\newcommand{\Sstep}{\constf{S}_\textsf{step}}
\newcommand{\measurable}{\mathbin{\rightarrow_M}}
\newcommand{\bindop}{\mathbin{>\!\!\!>\mkern-6.7mu=}}
\newcommand{\tendsto}{\xrightarrow{\hphantom{AAA}}}

\const{\bind}{bind}

\const{\MDPreward}{MDP-reward}
\const{\MDP}{MDP}
\const{\C}{C}

\const{\Ane}{A-ne}
\const{\subprobalgebra}{subprob-algebra}
\const{\policystep}{policy-step}
\const{\policyimprovement}{policy-improvement}
\const{\policyiteration}{policy-iteration}
\const{\streamspace}{stream-space}
\const{\completespace}{complete-space}
\const{\argmaxA}{arg-max}
\const{\hasargmax}{has-arg-max}
\const{\maxLex}{max-L-ex}
\const{\findpolicy}{find-policy}
\const{\vipolicy}{vi-policy}
\const{\conserving}{conserving}
\const{\clog}{log}
\const{\vi}{value-iteration}
\const{\improving}{improving}
\const{\pmf}{pmf}
\const{\isdec}{is-dec}
\const{\ispolicy}{is-policy}
\const{\isdecdet}{is-dec-det}
\const{\mkdecdet}{mk-dec-det}
\const{\mkstationary}{mk-stationary}
\const{\falseA}{False}
\const{\trueA}{True}
\const{\Suc}{Suc}
\const{\fst}{fst}
\const{\snd}{snd}
\const{\probspace}{$\mathcal{P}$}
\const{\tracespace}{$\mathcal{T}$}

\const{\return}{return}
\const{\cempty}{empty}

\const{\prob}{$\mathbb{P}$}
\const{\setpmf}{set-pmf}
\const{\mappmf}{map-pmf}
\const{\mapA}{map}
\const{\returnpmf}{return-pmf}
\const{\cspace}{space}
\const{\sets}{sets}
\const{\indicator}{indicator}
\const{\countspace}{count-space}
\const{\clim}{lim}
\const{\csup}{$\bigsqcup$}
\const{\policies}{$\Pi$}
\const{\vecB}{$V_B$}

\const{\bfun}{bfun}
\const{\bounded}{bounded}
\const{\range}{range}
\const{\undefined}{undefined}
\const{\dist}{$d_\infty$}
\const{\reverse}{reverse}
\const{\IT}{IT}
\const{\norm}{norm}
\const{\borel}{borel}
\const{\T}{T}
\const{\distr}{distr}
\const{\univ}{UNIV}
\const{\real}{$\mathbb{R}$}
\const{\ereal}{ereal}
\const{\Up}{Up}
\const{\Right}{Right}
\const{\Down}{Down}
\const{\action}{action}
\const{\Left}{Left}
\const{\cAE}{AE}
\const{\cin}{in}
\const{\X}{X}
\const{\Y}{Y}
\const{\Pt}{$\mathcal{X}$}
\const{\id}{id}
\const{\Trap}{Trap}
\const{\stateA}{state}
\const{\Pos}{Pos}
\newcommand{\LL}{\mathcal{L}}
\newcommand{\LLb}{\mathcal{L}_b}
\newcommand{\Kst}{\textsf{K}_\mathsf{st}}
\newcommand{\EK}{\mathcal{K}_\mathsf{st}}
\newcommand{\pushexp}{\textsf{pushexp}}
\newcommand{\rdec}{\textsf{r}_\mathsf{dec}}
\newcommand{\rb}{\textsf{r}_\mathsf{b}}
\const{\etr}{etr}
\newcommand{\etrfin}{\textsf{etr}_\mathsf{fin}}
\newcommand{\etropt}{\nu^*}
\newcommand{\rM}{\mathsf{r}_\mathsf{M}}
\newcommand{\TT}{\mathcal{T}}
\newcommand{\XX}{\textsf{X}}
\newcommand{\YY}{\textsf{Y}}
\newcommand{\KKzero}{\textsf{P}_\textsf{X}}
\const{\condpmf}{cond-pmf}
\const{\asmarkovian}{as-markovian}
\const{\actstar}{act*}
\const{\finite}{finite}
\const{\isargmax}{is-arg-max}

\newcommand{\pisuc}{\pi\textsf{-Suc}}
\newcommand{\YX}{\textsf{Y}^\textsf{X}}

\key{\klocale}{locale}
\key{\typedef}{typedef}
\key{\datatype}{datatype}
\key{\kfix}{fixes}
\const{\cfix}{fix}
\key{\kand}{and}
\key{\klet}{let}
\key{\kin}{in}
\key{\kif}{if}
\key{\then}{then}
\key{\kelse}{else}
\key{\kdo}{do}
\key{\kcase}{case}
\key{\kof}{of}
\key{\assume}{assumes}
\key{\kshow}{shows}

\type{\boolA}{bool}
\type{\nat}{$\mathbb{N}$}
\type{\set}{set}
\type{\tlist}{list}
\type{\stream}{stream}
\type{\measurepmf}{measure-pmf}
\type{\probalgebra}{prob-algebra}
\type{\vsigma}{vsigma}
\type{\streams}{streams}
\type{\cbind}{bind}
\type{\measure}{measure}
\type{\emeasure}{emeasure}
\type{\metricspace}{metric-space}
\type{\realnormedvector}{real-normed-vector}
\type{\realvector}{real-vector}
\type{\countable}{countable}
\type{\dec}{dec}
\type{\pol}{pol}

 \section{Introduction}
\label{sec:intro}

Despite the impressive advances in the capabilities of different types of AI systems, it is becoming clear that one major hurdle to their wide adoption is the lack of trustworthiness of these systems.
This has prompted researchers to study techniques to boost the trustworthiness of AI systems in different areas, like machine learning~\cite{SelsamNNVerification,katzReluplexEfficientSMT2017}, planning~\cite{DBLP:conf/itp/AbdulazizGN19,ictai2018}, and model-checking~\cite{esparza2013fully}.
However, one of the areas of AI where there is still a lot to be done regarding trustworthiness is software for solving Markov decision processes~(MDPs).
MDPs are models for systems where a decision-maker selects actions with random outcomes
to maximize long-term rewards.
Such systems with uncertainty occur in wide-ranging areas, e.g.\ planning, reinforcement learning, model checking, and operations research.
Depending on the area, the questions one asks about the model might be different, e.g.\ in planning the aim might be to find a policy that chooses an action for a robot in every state, with some optimality guarantees on the accrued rewards, while in model-checking the aim might be, for instance, to find what the expected delays of a real-time system are.
Furthermore, a lot of safety-critical systems could be modeled as MDPs, e.g.\ a policy (aka strategy) could be used to navigate an autonomous vehicle.
Like with other AI systems, for such applications, it is strongly desirable to have trustworthy programs to reason about and solve MDPs.

A methodology with notable success in developing provably correct software involves the use of interactive theorem provers~(ITPs).
Success stories of the use of ITPs to develop trustworthy software include a formally verified OS kernel~\cite{klein2009sel4}, a verified compiler for C~\cite{leroy2009formal}, and verified implementations of a multitude of algorithms~\cite{nipkowVerifiedTextbookAlgorithms2020}.

In this work, we study the application of ITPs to the development of trustworthy software for reasoning about and solving MDPs using iterative methods.
However, using ITPs to develop verified implementations of these iterative methods brings its own particular set of challenges, compared to developing other types of verified algorithms.
First, at a mathematical level, formal proofs of correctness of MDP solving algorithms in ITPs need a combination of diverse and significantly non-trivial formal mathematical libraries and concepts, e.g.\ probabilities, limits, (co)recursion, and probabilistic transition systems.
Second, at an implementation level, multiple challenges exist, e.g.\ should the algorithm be implemented imperatively or functionally, what kind of implementation of numerics should be used, etc.
Last, an even bigger challenge is the overall architecture of the verified system: should it be a verified implementation of an iterative method, or should we use an unverified system to perform the computation and produce a certificate, which is later validated using a formally verified certificate checker.

In addressing those challenges, multiple factors play in, like the feasibility of the verification, the reusability of the results, and the efficiency or practicality of the verified implementation.
Although previous authors~\cite{VajjhaSTPF21,DBLP:journals/corr/abs-2112-05996} have addressed the problem of verifying iterative methods for MDPs using ITPs, all of them focused on the abstract mathematical challenges.

In this paper, we comprehensively tackle all three challenges.
At the abstract mathematical level, using the ITP Isabelle/HOL~\cite{DBLP:books/sp/NipkowPW02} and following the exposition of Puterman~\cite{Puterman94}, we develop a general formalization of MDPs with rewards.
Based on that, we first formally prove correct the backward induction algorithm for finite-horizon MDPs.
Second, for infinite-horizon problems, we model four fundamental iteration-based methods in Isabelle/HOL, namely, value iteration, policy iteration, modified policy iteration, and splitting-based methods.
We prove the correctness of all methods against a formal specification of expected total discounted reward.
Compared to previous attempts~\cite{VajjhaSTPF21,DBLP:journals/corr/abs-2112-05996}, at a mathematical level, our formalization is more reusable. 
We also show the correctness of the algorithms against a simpler definition of expected total discounted reward.

As our second contribution, we devise the first executable verified implementations of the four algorithms of which we are aware.
In the process, we fix a mistake in a textbook correctness proof of Gauss-Seidel value iteration.
We experimentally evaluate our implementations of the four algorithms on standard probabilistic planning problems and show that they are practical.
Our implementations use efficient data structures and can solve planning problems with millions of transitions. 

Finally, we experimentally show that combining our verified infinite-horizon implementations with an unverified implementation yields significant performance improvements.
In particular, we show that one can use a fast floating-point implementation to perform all the iterations and then use the formally verified implementation for the last iteration, effectively having the best of both worlds.
The formalization, our verified implementation, and detailed benchmarks are available online.\footnote{https://github.com/schaeffm/mdps-isabelle-hol}

 \section{Background}
\label{sec:bg}

\subsection{Isabelle/HOL}

An ITP is a program that implements a formal mathematical system, in which definitions and theorem statements are written, and a set of axioms or derivation rules, using which proofs are constructed.
To prove a fact in an ITP, the user provides high-level steps of a proof, and the ITP fills in the details, at the level of axioms, culminating in a formal proof.
We performed the formalization in this paper using the interactive theorem prover Isabelle/HOL~\cite{DBLP:books/sp/NipkowPW02}, a theorem prover for Higher-Order Logic.
Roughly speaking, Higher-Order Logic can be seen as a combination of functional programming with logic.
Isabelle/HOL supports the extraction of the functional fragment to executable code~\cite{haftmann2007code}.

Isabelle is designed for trustworthiness: following the Logic for Computable Functions (LCF) approach~\cite{milner1972logic}, a small kernel implements the inference rules of the logic, and, using encapsulation features of abstract data types, it guarantees that all theorems are actually proved by this small kernel.
Around the kernel, there is a large set of tools that implement proof tactics and high-level concepts like algebraic data types and recursive functions.
Bugs in these tools cannot lead to invalid proofs, but only to error messages when the kernel refuses a proof.

\subsection{Probability Theory}
Probability theory was formalized in Isabelle/HOL primarily by Hölzl based on a library of measure theory~\cite{Holzl13}.
Let $\Omega$ be the sample space of the probability space $P$.
We denote the expectation of a random variable $X : \Omega \to \real$ by $\mathbb{E}_{\omega \sim P} \left[ X(\omega) \right]$, it is defined via the Lebesgue integral.
We write $\probspace(\Omega)$ for the set of all probability spaces over $\Omega$.

\subsection{Bounded Functions}
\label{sub:bounded_functions}

Our formal analysis of MDPs makes use of bounded functions.
For a set $X$ and a complete normed vector space $V$, 
the space of bounded functions $X \to_b V$ contains all functions where the image of $X$ is bounded w.r.t. the norm on $V$.
We define a pointwise addition and scaling operation, and a partial order for bounded functions.
The uniform norm makes bounded functions a complete normed vector space.

The set of bounded linear functions $V \to_L W$ between normed vector spaces consists of all linear transformations that perform bounded scaling.
For finite-dimensional vector spaces, bounded linear functions correspond to matrices.
Equipped with the operator norm $\| A \| \coloneqq \sup_{v \in V} \| Av \| / \| v \|$, bounded linear functions also form a normed vector space.
We formalize the following closed-form formula for the geometric series of contractive bounded linear functions, that is central to our analysis:

\begin{theorem}[{\citeauthor{Puterman94}~\citeyear[Corollary C.4]{Puterman94}}]\label{thm:geom_matrix}
  Let $A: V \to_L W$ with $\| A \| < 1$, then
  $\sum_{i \in \nat} A^i = (1 - A)^{-1}$.
\end{theorem}

Bounded linear functions are available in the Isabelle/HOL distribution as a subtype of the function type.
A type for bounded continuous functions is also present in the library, which we generalize to bounded functions.
Introducing special function types has the advantage that we can then show that they are instances of several vector space type classes~\cite{HaftmannW06}
and profit from an existing library of theorems and overloaded notation for vector spaces.

 \section{Markov Decision Processes}

In this section we give an overview of our formalization of the mathematical concepts needed for specifying the different algorithms on MDPs and their correctness criteria.
We closely follow the exposition in \cite[Chapters~2-6]{Puterman94}.
However, the most notable outcome of our formalization is that we give a more explicit construction of the trace space of the MDP and we also correct mistakes in the proofs.

Markov decision processes \cite{Bel57} model systems where an agent acts in an environment that exhibits randomized behavior.
The dynamics of MDPs evolve over a succession of epochs,
during each of which the agent analyzes the current state of the environment, chooses an action, obtains a reward, and transitions to a new state.
The goal of the agent is to optimize the action selection to maximize the accrued rewards.
We distinguish between finite and infinite horizon problems, where the number of epochs is finite or infinite respectively. 
Our formalization defines MDPs on general state and action spaces, but our analysis of solution methods only covers discrete state and action spaces.

\begin{definition}[Discrete Markov Decision Process with Rewards]
    A Markov decision process is composed of
        discrete sets of states $S$ and actions $A$, 
        a Markov kernel of transition probabilities $K : S \times A \to \probspace(S)$, 
        a bounded reward function $r : S \times A \to_b \real$,
        and state rewards $r_N : S \to_b \real$.
        For each state-action pair, $r$ gives a real-valued reward,
        while $r_N$ denotes the final value of a state (for finite-horizon problems). 
        Furthermore, we assume the existence of a non-empty set of enabled actions $A_s \subseteq A$ for each state $s \in S$.
\end{definition}

We use \emph{locales} \cite{LocalesBallarin} to define MDPs\listingref{ (Listing~\ref{snip:locale})} in Isabelle/HOL.
A locale can be seen as a formal mathematical context that introduces constants and assumptions, in which we develop our formalization.
Locales can be instantiated, e.g. with a concrete MDP.
This requires a proof that discharges the assumptions, yielding all the theorems proved within the locale. 
There already exists a formalization of MDPs in the Archive of Formal Proofs \cite{Holzl17}.
However, this formalization does not support stochastic action choice, an important component of our analysis.
Hence we develop a more flexible definition of MDPs ourselves.

\subsection{Construction of the Trace Space}
In each epoch, the agent determines the action with a decision rule $d : S \to \probspace(A)$, 
i.e. it chooses a probability distribution over the enabled actions.
A deterministic decision rule is a function $d : S \to A$ that respects enabled actions.
In general, the decision rule employed by the agent may depend upon the history of all previous decisions and states visited.
Histories are alternating sequences of states and actions, they form the set $H$.
A policy $\pi : H \times S \to \probspace(A)$ can be seen as a mapping from histories to decision rules.
We denote the set of all decision rules by $D$ and the set of policies by $\Pi$.
Markovian policies depend only on the current epoch and may thus be represented as a sequence of decision rules.
Stationary policies are policies that only use a single decision rule.
In our formalization, each subclass of policies has a different type.
We insert explicit coercion functions to translate between different types of policies,
but we omit them here for notational economy.

We now fix a policy $\pi$ and an initial state. 
That determines the trace space of the MDP,
the probability space of infinite sequences of state-action pairs as observed by the agent.
The law of this stochastic process is $\kappa :H \rightarrow \probspace(S \times A)$\listingref{ (Listing~\ref{snip:law})}:
\begin{alignat}{6}
     & \kappa(s) && \coloneqq \kdo \{ ~&& a \leftarrow \pi(s); ~ \returnop\ (s,a) ~ \} \\
     & \kappa(h, s, a) && \coloneqq \kdo \{~ &&s' \leftarrow K(s, a); ~
    a' \leftarrow \pi(h, s, a, s');\nonumber\\
    &&&&&\returnop\ (s',a') ~\}.\span\span\span
\end{alignat}

The definition uses the Giry monad \cite{Giry80} on probability spaces to elegantly compose a sequence of experiments.
A more detailed introduction to the Giry monad and \textbf{do}-notation is given in the appendix.
Specifically, the law $\kappa$ shows how to extend a finite trace by a single state-action pair:
we sample the next state $s'$ using the Markov kernel $K$ on the current state-action pair from the history,
then choose an action, 
and finally return the next state-action pair.
From the law of the stochastic process, we obtain the trace space $\tracespace^\pi : S \to \probspace((S \times A)^\infty)$ via the Ionescu-Tulcea extension theorem. 
The theorem was formalized by Hölzl, who used it to construct trace spaces for Markov chains~\cite{Holzl17}.

Alternatively, the stochastic process can be viewed as a sequence of state-action distributions, one for each epoch.
The state-action distribution $\cP^n_\pi : S \to \probspace(S \times A)$ at epoch $n$ is obtained as a projection from $\tracespace^\pi$\listingref{ (Listing~\ref{snip:pn})}.
We prove that $\cP^n_\pi$ adheres to the following recursive characterization:
\begin{alignat}{7}
     & \cP^{0}_\pi(s) &  & = \kdo \{~ a \leftarrow \pi(s); ~ \returnop\ (s,a) ~\}                                 \\
     & \cP^{n+1}_\pi(s) &  & = \kdo \{~ a \leftarrow \pi(s); ~ s' \leftarrow K(s, a); ~ \cP^{n}_{\pi'}(s') ~ \} \\
     & \quad \text{where}~\pi'((s_1,a_1),\dots,t) = \pi((s, a),(s_1,a_1),\dots,t).\span\span\span\span\span\nonumber
\end{alignat}

From these equations we derive that for a fixed initial state $s \in S$ and an arbitrary policy $\pi$, there exists a Markovian policy $\pi_M$ such that $\cP^n_\pi(s) = \cP^n_{\pi_M}(s)$ for all $n$\listingref{ (Listing~\ref{snip:sim})}.
This fact allows us to restrict the search for optimal infinite-horizon policies to Markovian policies.

\subsection{Finite-Horizon MDPs}

For a finite horizon $N$, we denote the value of a policy $\pi$ as $\nu^\pi_N : S \to_b \real$. 
It is defined as the expected discounted sum of rewards accrued over time, plus a final state reward:
    \begin{align}
        &\nu^\pi_N(s) \coloneqq \mathbb{E}_{\omega \sim \tracespace^\pi(s)} \left[ \textstyle\sum_{i < N} \lambda^i r(\omega_i) + r_N(\omega_{N,1}) \right].
    \end{align}

Typically for finite-horizon problems the discount factor $\lambda = 1$.
We show that the optimal finite-horizon value
$\nu_N^* \coloneqq \sup_{\pi \in \Pi} \nu_N^\pi$ can be computed
using backward induction as $\nu^*_N = u_0^*$, where $u_N^*(s) \coloneqq r_N(s)$ and for $t < N$
$$
\textstyle u_t^*(s)\coloneqq \sup_{a \in A_s} r(s, a) + \lambda \mathbb{E}_{K(s,a)} \left[ u^*_{t+1}\right].
$$

The maximizing actions in each equation constitute an optimal, Markovian, and deterministic policy.

\subsection{Infinite-Horizon MDPs}
We also define the infinite-horizon value $\nu^\pi : S \to_b \real$ 
    \begin{equation}
    \label{def:etr}
        \nu^\pi(s) \coloneqq \lim_{n \to \infty} \nu^\pi_n(s).
    \end{equation}

For infinite-horizon problems we assume $0 \le \lambda < 1$, as well as $r_N = 0$.
The discount factor decreases the relevance of later rewards, and also guarantees that each policy has a finite value.
We show that $\nu^\pi$ is a member of the space $V_B$ of bounded functions $S \to_b \real$. 
Here, we would like to note that both existing formalizations of the expected total discounted reward criterion~\cite{DBLP:journals/corr/abs-2112-05996,VajjhaSTPF21} avoid the construction of the trace space and, instead, use the following characterization as their definition of $\nu^\pi$:
\begin{align}\label{eq:etr_eq}
     \nu^\pi(s) = \textstyle\sum_{i \in \nat} \lambda^i\ \mathbb{E}_{\cP^i_\pi(s)} \left[r\right].
\end{align}

We argue that our definition is more natural and simpler, as it is derived from the trace space of the MDP, and is the one used in the textbook exposition, e.g.~\citeauthor{Puterman94}~\citeyear[Equation~5.1.3]{Puterman94}.
Thus, we believe it is a superior basis for the verification of specifications in terms of $\nu^\pi$.
However, the definition of $\nu^\pi$ is difficult to work with directly, because reasoning w.r.t.\ the trace space is complex.
We therefore show that our definition and (\ref{eq:etr_eq}) are equivalent.

We further simplify $\nu^\pi$ with the introduction of a vector notation.
The reward vector for a decision rule $d$ is defined as $r^d_s \coloneqq \mathbb{E}_{a \sim d(s)} \left[ r(s, a) \right].$
We also define the operator $\PP^n_\pi : \vecB \to_L \vecB$, a bounded linear function that is equivalent to the $n$-step transition probability matrix for policy $\pi$.
$\PP^n_\pi$ acts in the same way on a function as a stochastic matrix does on a vector.
We prefer statements in vector notation in our formalization, so we can prove theorems at a high level of abstraction.
Now we can formulate $\nu^\pi$ in vector notation for Markovian policies $\pi$\listingref{ (Listing~\ref{snip:vector})}:
\begin{align}
     &\nu^\pi = \textstyle\sum_{i} \lambda^i \PP^i_\pi r^{\pi_i},\ (\PP_\pi^iv)_s \coloneqq \mathbb{E}_{(s', a') \sim \cP^i_\pi(s)} \left[v_{s'}\right].
\end{align}

Ultimately our goal is to maximize the expected total discounted reward by optimizing the policy.
We define the value of the MDP
$\nu^* \coloneqq \sup_{\pi \in \Pi} ~ \nu^\pi$\listingref{ (Listing~\ref{snip:etropt})} to be the least upper bound of $\nu^\pi$ ranging over all policies $\pi$.
As noted earlier, it suffices to consider Markovian policies for optimality.

\subsubsection{Policy Evaluation}
Before searching for an optimal policy, we first tackle the problem of policy evaluation, i.e. the computation of $\nu^\pi$.
Specifically, we only consider a decision rule $d$.
It is possible to compute $\nu^d$ exactly, as from Theorem~\ref{thm:geom_matrix} we can show that $\nu^d = (1 - \lambda \PP_d)^{-1} r^d$\listingref{ (Listing~\ref{snip:inv})}.
However, determining an exact solution becomes computationally expensive for large state spaces.
A more practical alternative is the approximation using a fixed-point iteration.
We define the Bellman operator $\cL_d(v) \coloneqq r^d + \lambda \cP_d v$ and observe that
$\nu^d = \cL_d(\nu^d)$.
Since $\cL_d$ is a contraction mapping, the Banach fixed-point theorem establishes that $\nu^d$ is its unique fixed-point and that
$\lim_{i \to \infty} \cL_d^i(v) = \nu^d$ for any $v \in \vecB$.

\subsubsection{Optimality Equations}

The Bellman operator can be used to approximate $\etropt$.
We therefore introduce the Bellman optimality operator $\LL : \vecB \to \vecB$\listingref{ (Listing~\ref{snip:bellmanopt})} where
\begin{equation}\textstyle
\LL(v) \coloneqq \sup_{d \in D} ~ \cL_d(v).
\end{equation}

The proof that $\LL$ is well-defined is missing from~\citeauthor{Puterman94}'s book, as discussed by~\citeauthor{DBLP:journals/corr/abs-2112-05996}~\citeyear{DBLP:journals/corr/abs-2112-05996}.
We can prove that the supremum is indeed well-defined since changing the action selection for one state does not influence the value of other states.
Next, we prove\listingref{ (Listing~\ref{snip:sols})} that every fixed-point of $\LL$ is equal to the optimal value $\etropt$.
Since $\LL$ is also a contraction mapping, $\etropt$ is actually the unique fixed-point of $\LL$ and
can be computed with a fixed-point iteration.

\begin{theorem}[Fixed point of $\LL$]
    \label{thm:bellman_opt_conv} The optimal value $\etropt$ is the unique fixed point of the optimality operator $\LL$. Additionally, $\lim_{i \to \infty} \LL^i(v) = \etropt$ for $v \in \vecB$.
\end{theorem}

It remains to be shown under which conditions an optimal policy (one that achieves the value of the MDP) exists.
A sufficient condition is the existence of the supremum in the definition of $\LL(v)$ for every $v \in \vecB$.
The optimal policy is then the decision rule $d^*$ that maximizes $\cL_{d^*}(\etropt)$\listingref{ (Listing~\ref{snip:ex})}.
If the set of enabled actions in each state is finite, the supremum in $\LL$ is always attained.
Thus finite MDPs have optimal policies that are stationary and deterministic.
Finally, we show that if there exists an optimal policy, there also exists an optimal stationary and deterministic policy.

\section{Infinite-Horizon Algorithms}
A novelty of our work is that we verify optimized and executable algorithms to compute optimal policies.
Verifying such executable algorithms pertains to two tasks.
The first is formalizing the algorithm in the logic of the theorem prover, usually as a non-executable recursive function.
This function should capture the mathematical essence of the algorithm, but excludes implementation details.
Then one proves the desired mathematical properties of this function, most notably termination and that the algorithms compute optimal policies.
In the second step, we formally verify executable versions of the algorithms at hand.
We now discuss the first step and then later discuss the second.

From here on, we assume that both $S$ and $A$ are finite.
This ensures the existence of optimal policies and allows us to extract executable code.
We shortly present the basic value iteration and policy iteration algorithms and then give a more detailed exposition of the optimized variants, namely modified policy iteration and Gauss-Seidel value iteration.
For finite-horizon MDPs, we compute optimal policies using the backward induction algorithm described earlier.

The value iteration algorithm\listingref{ (Algorithm~\ref{alg:vi}, Listing~\ref{snip:vi})} is based on Theorem~\ref{thm:bellman_opt_conv}. 
It repeatedly applies the Bellman optimality operator $\LL$ to an initial estimate of the value function and
stops as soon as successive iterates are sufficiently close in distance to achieve the desired accuracy.
On the other hand, the policy iteration algorithm\listingref{ (Algorithm~\ref{alg:pi}, Listing~\ref{snip:pi})} performs a direct search in the space of deterministic decision rules~\cite[Section~6.4]{Puterman94}.
It alternates between evaluating a candidate policy and improving it, until the policy stabilizes.
Exact policy evaluation involves solving a system of linear equations, 
for which we use a verified implementation of the Gauss-Jordan algorithm~\cite{thiemannJordanNormalForms}.

\paragraph*{Modified Policy Iteration}
Value iteration and policy iteration can be considered extreme instances of 
modified policy iteration (Algorithm~\ref{alg:mpi}\listingref{, Listing~\ref{snip:mpi}}) \cite[Section~6.5]{Puterman94}.
In each step, value iteration only approximates the value of the current policy with a single iteration of the Bellman operator, while policy iteration considers infinitely many iterations.
Modified policy iteration places a varying amount of policy evaluation steps between each policy improvement phase.
When the initial value estimate $v$ is conservative, i.e. $v \le \LL(v)$, the algorithm terminates with a policy that is optimal upto an a priori error bound ($\epsilon$-optimal policy).
The convergence proof of modified policy iteration is based on the observation that its iterates 
dominate the iterates of value iteration but never exceed $\etropt$.

\paragraph*{Gauss-Seidel Value Iteration}
Finally, Gauss-Seidel value iteration (Algorithm \ref{alg:vigs}\listingref{, Listing~\ref{snip:gs}}) is a variant of value iteration where the value estimates are updated in-place \cite[Section 6.3.3]{Puterman94}.
The algorithm delivers an $\epsilon$-optimal policy. It converges at least as fast as value iteration, because updated estimates are used immediately, not only in the next iteration.
In practice, often substantially fewer iterations are required until convergence.
We now assume that the state space of the MDP is a subset $\{0,1,\ldots,n\}$ of the natural numbers and each iteration proceeds from low to high states.
Thus we can interpret bounded linear functions as matrices.

\citeauthor{Puterman94} shows that Gauss-Seidel value iteration is an instance of a class of algorithms called splitting methods\listingref{ (Listing~\ref{snip:splitting})}.
These use regular splittings $(Q_d, R_d)$ of the linear function $(1 - \lambda \cP_d) = Q_d - R_d$.
Regularity of a splitting means that $Q_d^{-1}$ and $R_d$ are nonnegative matrices.
The algorithm proceeds the same way as value iteration with the Bellman operator replaced by
$\GS_d(v) \coloneqq Q_d^{-1} (r^d + R_d v)$.
For the Gauss-Seidel method we split $\cP_d = \cP^L_d + \cP^U_d$ into a strictly lower triangular part $\cP^L_d$ and an upper triangular part $\cP^U_d$ and
choose the regular splitting $Q_d = (1 - \lambda\cP^L_d)$ and $R_d = \lambda\cP^U_d$.
In-place value iteration should follow the equation $v^{n+1} = r^d + \lambda \cP^L_d v^{n+1} + \lambda\cP^U_d v^n$ for some $d$.
Rearranging terms then directly yields 
\begin{equation}
v^{n+1} = (1 - \lambda \cP^L_d)^{-1}(r^d + \lambda \cP^U_d v^n) = \GS_d(v^n).
\end{equation}

We define the Gauss-Seidel variant of the Bellman optimality operator $\mathcal{G}(v) \coloneqq \sup_{d \in D} \GS_d(v)$.
We show that for any regular splitting, 
$\mathcal{G}$ is a contraction mapping and the algorithm converges
if the supremum in $\mathcal{G}(v)$ is attained for all $v\in\vecB$ and $\sup_{d \in D}  \| Q_d^{-1}R_d\| < 1$. 
The requirement that the supremum has to be attained is overlooked by~\citeauthor{Puterman94}, where the existence of such a decision rule is implicitly assumed.
We close this proof gap for Gauss-Seidel value iteration.
Assuming a total ordering on the states, as part of our formalization, we show that a decision rule that attains the supremum for all states smaller than $s$ can always be extended to an optimal decision rule up to and including state~$s$.
Before we state the theorem, let $f (x:=y)$ denote the pointwise update of the value of function $f$ at $x$.

\begin{theorem} Assume that: 
    \begin{enumerate}
        \item the decision rule $d^*$ maximizes $\GS_{d^*}(v)$ for all states smaller than $s$, i.e.\ $(\GS_d(v))_{t} \le (\GS_{d^*}(v))_{t}$, for any $t < s$ and $d \in D_D$, and 
        \item the maximizing action in $s$ is $a$, i.e. for any $a' \in A_s$, $(\GS_{d^*(s := a')}(v))_s \le (\GS_{d^*(s := a)}(v))_s$.
    \end{enumerate}
Then $d^*(s := a)$ maximizes $\GS_{d^*}(v)$ for all states up to $s$, i.e.\ $(\GS_d(v))_{t} \le (\GS_{d^*(s:=a)}(v))_{t}$, for any $t \le s$ and $d \in D_D$.
\end{theorem}
\proof{
We obtain $\GS_d(v) = r^d + \lambda\cP^U_d v + \lambda\cP^L_d \GS_d(v)$ by rearranging the definition of $\GS_d$.
Since $\cP^L_d$ is lower diagonal, the value of $(\GS_d(v))_t$ depends only on the values of $d$ up to state $t$.
Now, it follows from the assumptions that $d^*(s:=a)$ is already maximizing for all $t < s$.
Thus it suffices to show that for all deterministic decision rules $d$, $(\GS_d(v))_s \le (\GS_{d^*(s:=a)}(v))_s$.
With $d' \coloneqq d^*(s := d(s))$, we have
\begin{align}
    & (\cP^L_d~\GS_d(v))_s\\  ={} &(\cP^L_{d'}\GS_d(v))_s      && \text{$\cP^L$ only depends on $d$ at $s$}\\
                         \le{}& (\cP^L_{d'}\GS_{d^*}(v))_s && \text{$d^*$ optimal, $\cP^L$ is triangular}\\
                         \le{}& (\cP^L_{d'}\GS_{d'}(v))_s. && \text{$\GS_{d^*}$ independent of $d^*$ above $s$}
\end{align}
So
$(\GS_d(v))_s = (r^{d} + \lambda \cP^U_{d}v + \lambda \cP^L_{d}\GS_{d}(v))_s \le (\GS_{d'}(v))_s.$
Finally assumption~2 lets us complete the proof.\qed
}

In Puterman's book, a proof of the $\epsilon$-optimality of this splitting-based method is not given, which we supplement.
However, that requires us to change the last step of the algorithm from how it was presented originally~\cite[Section 6.3.3]{Puterman94}. There, the policy is determined using a basic value iteration step.
To prove $\epsilon$-optimality, we need to, instead, use a Gauss-Seidel step to determine the policy.

\SetAlgoSkip{}
\begin{algorithm}
    \SetKwData{Left}{left}\SetKwData{This}{this}\SetKwData{Up}{up}
    \SetKwFunction{Union}{Union}\SetKwFunction{FindCompress}{FindCompress}
    \SetKwInOut{Input}{Input}\SetKwInOut{Output}{Output}
    \SetKwFor{Repeats}{repeat}{times}{}
    \SetKwFor{RepeatInf}{for}{do}{}

    \Input{$v \in \vecB$ where $v \le \LL(v)$, $m : \mathbb{N} \to \mathbb{N}$}
    \DontPrintSemicolon
    \RepeatInf(){$i \in 0\ldots$}{
        \lFor{$s \in S$}{
            $d(s) \gets \argmax_{a \in A_s} (\cL_a(v))_s$
        }

        \lIf{$\dist(v, \LL(v)) < \frac{\epsilon (1 - \lambda)}{2\lambda}$}{\Return{d}}

        $v \gets \cL_d^{m_i + 1}(v)$\\
    }

    \caption{Modified Policy Iteration}
    \label{alg:mpi}
\end{algorithm}
\begin{algorithm}
   \SetKwData{Left}{left}\SetKwData{This}{this}\SetKwData{Up}{up}
   \SetKwFunction{Union}{Union}\SetKwFunction{FindCompress}{FindCompress}
   \SetKwInOut{Input}{Input}\SetKwInOut{Output}{Output}
    \SetKwFor{RepeatInf}{for}{do}{}
   \Input{$v \in \vecB$}

    \DontPrintSemicolon
   \Repeat(){$\dist(v, v_{old}) < \frac{\epsilon (1 - \lambda)}{2\lambda}$}{
      $v_{old} \gets v$\\
      \lFor{$s \in S$}{
          $v(s) \gets \max_{a \in A_s} (\cL_a(v))_s$}
   }
   \RepeatInf{$s \in S$}{
      $d(s) \gets \argmax_{a \in A_s} (\cL_a(v))_s$\\
      $v(s) \gets \max_{a \in A_s} (\cL_a(v))_s$
   }
   \Return{d}
   \caption{Gauss-Seidel Value Iteration}
   \label{alg:vigs}
\end{algorithm}

\section{Verifying an Executable Implementation}

Our approach to obtaining executable versions of the algorithms is based on step-wise refinement~\cite{refinementWirth}.
In this approach, an executable specification of an algorithm is derived from an unexecutable one by replacing mathematical structures by data structures.
We make heavy use of locales in our refinement.
The built-in data-refinement of the executable code generator of Isabelle/HOL proved to be too inflexible, as it does not allow the same mathematical structure to be implemented with different data structures at different places in the algorithm.

Initially, we have abstract definitions of the algorithms as presented above.
In the first step, we show that the reward function and the transition system can be represented as finite maps, and that action sets can be represented as finite sets.
The interfaces of data structures implementing finite sets and maps are available as locales in the Isabelle/HOL distribution.
These locales come with interpretations for concrete data structures based on e.g.\ balanced trees or hash maps.
We prove correctness on the level of the abstract interfaces, 
which gives us the flexibility to choose between several concrete data structures without modifying proofs.
In our final version, we represent an MDP as an array with an entry for each state. 
The array stores red-black trees that map actions to rewards and transitions.

We use the code generation facilities provided by Isabelle/HOL \cite{haftmann2007code} to extract executable Standard~ML code from the formally verified algorithms.
A wrapper parses problems from a file and passes them to the verified algorithms.
To obtain numerically accurate results, real numbers are represented as rational numbers.
This representation comes with an ever greater performance penalty as the number of iterations increases and the fractions become larger.
For comparison, we also provide a separate implementation using floating-point arithmetic.
However, due to the possibility of floating-point errors accumulating, we lose the formal guarantees.

\section{Notes on the Formalization}
Since a main goal of this project is to showcase theorem proving as a methodology to obtain verified algorithms for probabilistic systems, we note some of the lessons we learnt and the challenges encountered during the formalization.
This section should be of particular interest to readers who would be interested in verifying similar algorithms in Isabelle/HOL or in other theorem provers.

Our work builds on formalization efforts in probability theory \cite{Holzl13} and linear algebra from the archive of formal proofs~(AFP) and the Isabelle distribution.
The construction of MDPs and their trace space extends previous work in Isabelle/HOL on Markov Chains and MDPs \cite{Holzl17}.
This library of formalized mathematics was crucial to make our verification project viable.

Isabelle/HOL provides us with powerful tools for automated reasoning,
such as the proof method \emph{auto}, and access to external automated theorem provers via \emph{sledgehammer} \cite{paulson2010three}.
The strong automation combined with the structured proof language Isar~\cite{DBLP:conf/tphol/Wenzel99} enables the development of maintainable, reusable and human-readable proofs.
However, in our formalization we encounter situations where automation breaks.
For instance, chains of equations involving mathematical operators that are potentially undefined, like infinite sums or integrals require duplicate work.
We first need to show that the operation maintains the desired property, e.g. summability, measurability, boundedness and integrability.
Only then can we show that the algebraic manipulation is actually correct.
This problem becomes especially apparent with nested sums or integrals.
A potential, albeit still preliminary, solution to this problem has been proposed by~\cite{formalisingUndefinedTerms}.

To improve the reusability of our work, we formalize MDPs with arbitrary, uncountable state spaces and probabilistic action choice.
Still, the correctness of the verified algorithms only holds for finite-state MDPs.
Our initial formalization of the algorithms models the state space of the MDP as a type, i.e.\ all the states have to be known at compile-time.
Initially, our intention was to use the Isabelle tool \emph{types-to-sets} \cite{kuncarTypesToSets} to generalise our formalization to be parameterised by the set of states of the MDP, and get algorithms operating on the corresponding sets of states.
Nonetheless, we could not transfer statements involving arbitrary probability spaces automatically using that tool, as its automation is also still preliminary.
 \section{Experimental Evaluation}
We evaluate our algorithms on a set of standard benchmarks and compare the results to a reference implementation.
Furthermore, we show how our implementation can be supported with solutions obtained by unverified solvers.

\paragraph {Setup}
We benchmark our implementation on explicitly represented MDPs, which are compiled problems from the \emph{International Planning Competition 2018}.
\footnote{\url{https://ipc2018-probabilistic.bitbucket.io/}}
Most domains come with multiple instances of different sizes.
The problems are parsed by an unverified wrapper and are then handed to the verified solvers.
Each problem is given a timeout of four hours and a memory limit of 4~GB.
For infinite-horizon problems, we set a discount factor of $\lambda = 0.95$ and require an accuracy of $\epsilon = 0.05$ (for our verified policy iteration and Storm, $\epsilon = 0$).
Finite-horizon MDPs are run with $\lambda = 1$ and a horizon of $N = 50$.

We compare our work to the probabilistic model checkers PRISM~\cite{PRISM2006} and Storm~\cite{hensel2021probabilistic}, that we extended to support discount factors.
We use PRISM's explicit representation mode and its floating-point arithmetic mode, while we use Storm in its exact mode.
The point of this setup is to individually evaluate the performance impact of both precise arithmetic and verified data structures and algorithms.
For part of the benchmarks, we use the values generated by a hand-written implementation of Gauss-Seidel value iteration in Rust as initial values for the verified algorithms.
That way, the verified programs certify the unverified solutions to achieve better performance.

\paragraph{Results}

\begin{table*}
    \small
    \begin{tabularx}{0.9\linewidth}{l  c | c | c | c | c | c | c | c | c | c | c | c | c | c | c | c || c | c | c }
                                                                              &
        {\multirow{3}{*}{\rotatebox[origin=c]{90}{Instances}}}                &
        \multicolumn{3}{c|}{VI}                                               &
        \multicolumn{3}{c|}{GS}                                               &
        \multicolumn{3}{c|}{PI}                                               &
        \multicolumn{3}{c|}{MPI}                                              &
        \multicolumn{2}{c|}{Cert.}                                            &
         &
        \multicolumn{3}{c}{Fin-Horizon}                                                                                                                                                                                                  \\[4pt]
                                                                              &
                                                                              & \multicolumn{1}{c|}{\multirow{2}{*}{\rotatebox[origin=c]{90}{Prism}}} &
        \multicolumn{2}{c|}{Verified}                                           &
        \multicolumn{1}{c|}{\multirow{2}{*}{\rotatebox[origin=c]{90}{Prism}}} &
        \multicolumn{2}{c|}{Verified}                                           &
        \multicolumn{1}{c|}{\multirow{2}{*}{\rotatebox[origin=c]{90}{Prism}}} &
        \multicolumn{2}{c|}{Verified}                                           &
        \multicolumn{1}{c|}{\multirow{2}{*}{\rotatebox[origin=c]{90}{Prism}}} &
        \multicolumn{2}{c|}{Verified}                                           &
        \multicolumn{1}{c|}{\multirow{2}{*}{\rotatebox[origin=c]{90}{VI}}}    &
        \multicolumn{1}{c|}{\multirow{2}{*}{\rotatebox[origin=c]{90}{GS}}}    &
        \multicolumn{1}{c||}{\multirow{2}{*}{\rotatebox[origin=c]{90}{Storm}}} &
        \multicolumn{1}{c|}{\multirow{2}{*}{\rotatebox[origin=c]{90}{Prism}}} &
        \multicolumn{2}{c}{Verified}                                                                                                                                                                                                    \\[4pt]
                                                                              &                                                                       &
                                                                              & $\real$                                                               &
        $\mathbb{F}$                                                          &                                                                       &
        $\real$                                                               &
        $\mathbb{F}$                                                          &                                                                       &
        $\real$                                                               &
        $\mathbb{F}$                                                          &                                                                       &
        $\real$                                                               &
        $\mathbb{F}$                                                          &                                                                       & &   &    &
        $\real$                                                               &
        $\mathbb{F}$
        \\[5pt]
        academic-advising                                                             & 2                                                                     & -- & -- & 1 & -- & -- & 1 & -- & -- & -- & -- & 1  & 1 & 1 & 1  & -- & 1  & 1  & 1 \\
        crossing-traffic                                                               & 4                                                                     & 4  & 4  & 4 & 4  & 2  & 4 & 4  & 2  & 2  & 4  & 4  & 4 & 4 & 4  & 4 & 4 & 4  & 4 \\
        elevators                                                             & 8                                                                     & 5  & 2  & 6 & 5  & 1  & 6 & 5  & 2  & 2  & 5  & 2  & 6 & 6 & 6  & 5 & 7  & 2  & 6 \\
        game-of-life                                                          & 3                                                                     & 3  & -- & 3 & 3  & -- & 3 & 3  & -- & 3  & 3  & -- & 3 & 3 & -- & -- &3  & -- & 3 \\
        manufacturer                                                          & 2                                                                     & 2  & -- & 2 & 2  & -- & 2 & 2  & -- & 1  & 2  & 1  & 2 & 2 & 2  & 1 & 2  & 1  & 2 \\
        push-your-luck                                                                  & 5                                                                     & 5  & 5  & 5 & 5  & 5  & 5 & 5  & 2  & 2  & 5  & 5  & 5 & 5 & 5  & 5 &5  & 5  & 5 \\
        skill-teaching                                                        & 8                                                                     & 6  & 4  & 7 & 6  & 3  & 7 & 6  & 2  & 4  & 6  & 4  & 8 & 6 & 6  & 4 &6  & 4  & 6 \\
        triangle-tireworld                                                             & 6                                                                     & 4  & 4  & 4 & 4  & 4  & 4 & 4  & 2  & 2  & 4  & 4  & 4 & 4 & 4  & 4 & 6  & 4  & 4 \\
        wildfire                                                              & 1                                                                     & -- & -- & 1 & -- & -- & 1 & -- & -- & -- & -- & -- & 1 & 1 & 1  & -- & -- & -- & 1 \\
        wildlife-preserve                                                         & 8                                                                     & 6  & 5  & 8 & 6  & 4  & 8 & 6  & 4  & 6  & 6  & 6  & 8 & 8 & 8  & 6 & 8  & 6  & 8
\end{tabularx}

    \caption{\label{table:solved}
        A table showing the number of instances solved by different algorithms.
        The first column gives the number of instances per domain. Columns two to five show the performance of the PRISM implementations vs.\ our implementations of value, policy, Gauss-Seidel, and modified policy iteration, where $\real$ and $\mathbb{F}$ indicate precise and floating-point arithmetic respectively.
        The sixth column displays the results for Storm.
        The seventh column shows the performance of using an unverified implementation of value iteration followed by one last verified iteration.
        The last column shows the results for the finite-horizon case.}
\end{table*} 
We give an overview of the results in Table~\ref{table:solved}.
First, we observe that the verified implementations using floating-point arithmetic can often compete with PRISM.
We assume that this is in part due to PRISM being a generalist optimized to handle factored systems.
An exception here is policy iteration, because PRISM uses an approximate method to compute the value of the policy
that is much faster than our exact method based on Gaussian elimination.
Comparing our algorithms, we see that policy iteration is only viable for small problems, as the arithmetic complexity of Gaussian elimination grows cubically in the number of states.
The Storm model checker provides the same exact results but solves more instances. 
Thus certification of the Storm results is a promising direction for future work.  
Using floating-point arithmetic, the optimized algorithms consistently outperform value iteration.
When precise arithmetic is used, Gauss-Seidel value iteration becomes comparatively slow.
This is because the fractions representing the value estimates become larger more quickly for in-place updates, as they grow already over the course of a single iteration.

Overall, the experiments show a large performance gap between floating-point and precise arithmetic, especially as the number of iterations increases.
With each additional iteration, the precise representation of the current value estimates gets more complex.
This prompted us to combine the unverified floating-point Rust implementation for infinite-horizon problems with the verified precise arithmetic: the resulting values from the unverified solver are provided as initial values to the verified precise arithmetic implementation.
This way we get the best of both worlds: values that are formally guaranteed to be $\epsilon$-optimal, with performance characteristics comparable to unverified implementations\appendixenv{ (see Figure \ref{fig:sub1} in the appendix)}. \section{Conclusion and Discussion}

Our work provides a comprehensive formalization of the basics of MDPs with rewards in the interactive theorem prover Isabelle/HOL.
We then prove correct optimized algorithms to solve discounted MDPs. 
We show that it is feasible to solve non-trivial Markov decision processes with verified algorithms.
Overall, our formalization is comprised of approximately 13.000 lines of definitions and proofs.
One point worth highlighting is that we use the output of an unverified efficient implementation of value iteration as input to a verified implementation that uses precise arithmetic.
This leads to a system with formal guarantees with similar performance characteristics as an unverified system.
The results show the feasibility of verifying practical algorithms in the realm of reinforcement learning or probabilistic planning.
In addition to an extensive library of existing formal proofs, Isabelle/HOL provides powerful facilities for proof search, automation, and structuring that in combination allow the development of verified software.

\paragraph{Future Work} The most immediate extension of our work to handle much larger state spaces is to formalize data structures for factored representations of MDPs.
These representations are standard in both model-checking and planning.
Furthermore, in many applications, the current state of the environment is only partially known by the agent.
Such systems are modeled using partially observable MDPs.
Extending our work in this direction is an important step towards the verification of reinforcement learning algorithms.
Another interesting direction concerns how to handle arithmetic: instead of relying on floating-point arithmetic which does not account for the accumulation of errors, we might investigate the possibility of using interval arithmetic to provide error-bounds.
Lastly, verified Monte Carlo algorithms would allow us to deal with large state spaces.

\paragraph{Related Work}
There is a number of related verification approaches that have been tried in the general context of artificial intelligence.
The first such thread of research is based on using completely automated methods.
It has been applied in planning~\cite{eriksson2017unsolvability} and SAT~\cite{DBLP:conf/sat/WetzlerHH14}, and has recently received the most attention in the area of verifying robustness of neural networks~\cite{katzReluplexEfficientSMT2017}.
In this latter application, a specification of a given neural network's robustness is compiled into an SMT formula, which is then automatically proved/disproved by an SMT solver.
An advantage of this approach compared to using ITPs is that it is fully automated, i.e.\ one could prove a neural network safe without fully understanding it.
A disadvantage, however, is that SMT solvers are limited in terms of what specifications they can automatically solve.
This is most evident when trying to verify properties of algorithms or programs.
For instance, they cannot prove a specification stating that a given implementation of value iteration computes an optimal policy.

Another related approach is that of~\cite{SelsamNNVerification}~and~\cite{CoqNNAAAI2019}.
In that thread of work, the authors develop ITP frameworks to aid in the formal reasoning about machine learning models and neural network architectures.
Since they use ITPs, they are able to prove specifications that are more interesting than neural network robustness.
For instance, they are able to show that a certain learning algorithm can guarantee a certain generalisation error, which cannot be done using SMT solvers.

MDPs have been formalized in theorem provers before, e.g. to
analyze the semantics of pGCL \cite{Holzl17}, or verify soundness of off-policy evaluation \cite{DBLP:conf/itp/YeagerMNT22}.
The most closely related formalization is the project by \cite{VajjhaSTPF21} who develop the library \emph{CertRL} in the interactive theorem prover Coq.
Our work differs in a number of ways: they show correctness of value and policy iteration while we also prove correct more involved algorithms.    
Furthermore, they do not verify efficient implementations from their formalisations, while we do.
They also define the expected total discounted reward as equation~(\ref{eq:etr_eq}), while we give a simpler and more natural definition, which needs the construction of the trace space of an MDP.
We also show that deterministic decision rules are in fact optimal over all policies, while they only consider deterministic decision rules as candidates for optimal policies.
Furthermore, in Isabelle/HOL there exists a recent formalization of the basics of discrete reinforcement learning~\cite{DBLP:journals/corr/abs-2112-05996}.
Their approach uses stochastic matrices instead of the Giry Monad.
They also use equation (\ref{eq:etr_eq}) as their definition of the expected reward, only cover finite MDPs, do not discuss executable algorithms, and prove, at an abstract level, that value and policy iteration solve MDPs optimally.
 \section*{Acknowledgements}
This work was partially funded by the Deutsche Forschungsgemeinschaft Research Training Group CONVEY - 378803395/GRK2428 and the Deutsche Forschungsgemeinschaft Koselleck Grant NI 491/16-1.
We thank Thomas Keller for helpful comments and providing us with benchmark problems.
 \bibliography{short_paper}

\section{Appendix: Giry Monad}
The Giry Monad \cite{Giry80} allows to elegantly compose probability spaces in formal languages. 
It was formalized in Isabelle/HOL by \cite{EberlHN15}
and defines the usual two monad operations.
The function $\returnop : \Omega \to \probspace(\Omega)$ lifts an element of the sample space to a probability space.
For $x \in \Omega$, $\returnop(x)$ gives the Dirac measure at $x$, i.e. for an event $X$ we have
\begin{equation}
\mathbb{P}_{\returnop(x)}(X) = \kif ~ x \in X ~ \then ~ 1 ~ \kelse ~ 0.
\end{equation}

The second operator $\cbind$ (with infix notation $\bindop$) chains two random experiments, where the second experiment depends on the outcome of the first one.
Let $P \in \probspace(M)$, $Q : M \to \probspace(N)$, then $P \bindop Q \in \probspace(N)$.
For an event $X$ on $N$,
\begin{align}
  \prob_{P \bindop Q}(X) = \mathbb{E}_{x \sim P} \left[ \prob_{Q(x)}(X) \right].
\end{align}

Computations involving $\bindop$ and $\returnop$ are written in $\kdo$-notation as established by the functional programming language Haskell.
This notation can be desugared recursively as follows:
\begin{align*}
  & \kdo\{~x \leftarrow P ;~\textit{stmts}~\} \equiv P \bindop (\lambda x.~\textit{stmts}).
\end{align*}

\section{Appendix: Algorithms}

\SetInd{0.25em}{0.5em}
\DecMargin{0.3cm}
\begin{algorithm}
    \SetKwData{Left}{left}\SetKwData{This}{this}\SetKwData{Up}{up}
    \SetKwFunction{Union}{Union}\SetKwFunction{FindCompress}{FindCompress}
    \SetKwInOut{Input}{Input}\SetKwInOut{Output}{Output}
    \Input{$v \in \vecB$}

    \lWhile(){$d_\infty(v, \LL(v)) \ge \frac{\epsilon (1 - \lambda)}{2\lambda}$}{
        $v \gets \LL(v)$
    }
    \lFor(){$s \in S$}{
        $d(s) \gets \argmax_{a \in A_s} (\cL_a(v))_s$ }
    \Return{d}
    \caption{Value Iteration}
    \label{alg:vi}
\end{algorithm}

\begin{algorithm}
    \SetKwData{Left}{left}\SetKwData{This}{this}\SetKwData{Up}{up}
    \SetKwFunction{Union}{Union}\SetKwFunction{FindCompress}{FindCompress}
    \SetKwInOut{Input}{Input}\SetKwInOut{Output}{Output}
    \Input{$d \in D_D$}

    \Repeat(){$d = d_{old}$}{
        $d_{old} \gets d$\\
        $v \gets (1 - \lambda \cP_d)^{-1}r^d \qquad (= \nu^d)$\\
        \lFor{$s \in S$}{$d(s) \gets \argmax_{a \in A_s} (\cL_a(v))_s$}
        \qquad (if possible, keep $d$ unchanged)\\
    }
    \Return{d}
    \caption{Policy Iteration}
    \label{alg:pi}
\end{algorithm}

\section{Appendix: Isabelle/HOL Listings}
\todo{make this section more legible}

We relate the definitions and theorems from the paper to our formalization.
\vspace{1em}

\noindent
The function \texttt{K'} corresponds to $\kappa$ in the paper.
This is a more general version, that uses sequences instead of lists.

\begin{IsabelleSnippet}{Law of the Stochastic Process, label=snip:law}
definition K' :: "('s, 'a) pol \<Rightarrow> 's measure \<Rightarrow> nat \<Rightarrow> (nat \<Rightarrow> ('s \<times> 'a)) \<Rightarrow> ('s \<times> 'a) measure" where
  "K' p s0 n \<omega> = do {
    s \<leftarrow> case_nat s0 (K \<circ> \<omega>) n;
    a \<leftarrow> p n (\<omega>, s);
    return M (s, a)
  }"
\end{IsabelleSnippet}

\noindent Using the Ionescu-Tulcea extension theorem, we construct the trace space $\tracespace$ (\texttt{T} in the formalization).
\noindent The state-action distribution \texttt{Pn} corresponds to $P^n$ in the paper.
\begin{IsabelleSnippet}{State-Action Distributions, label=snip:pn}
primrec Pn :: "('s, 'a) pol \<Rightarrow> 's pmf \<Rightarrow> nat \<Rightarrow> ('s \<times> 'a) pmf" where
  "Pn p S0 0 = K0 (p []) S0" |
  "Pn p S0 (Suc n) = 
    K0 (p []) S0 \<bind> (\<lambda>sa. Pn (\<pi>_Suc p sa) (K sa) n)"

lemma Pn_eq_T: "measure_pmf (Pn p S0 n) = 
  distr (T p S0) (count_space UNIV) (\<lambda>t. t !! n)"
\end{IsabelleSnippet}

\noindent
We prove that for any initial state distribution \texttt{S0}, the policy \texttt{as\_markovian p S0} simulates \texttt{p}.
\begin{IsabelleSnippet}{Markovian Policies simulate History-Dependent Policies,label=snip:sim}
definition "Y_cond_X p S0 n x = 
  map_pmf snd (cond_pmf (Pn p S0 n) {(s,a). s = x})"

abbreviation "as_markovian p S0 n x \<equiv> 
  if x \<in> (Xn p S0 n) then Y_cond_X p S0 n x 
  else return_pmf (SOME a. a \<in> A x)"
  
theorem Pn_as_markovian_eq: 
  "Pn (mk_markovian (as_markovian p S0)) S0 = Pn p S0"
\end{IsabelleSnippet}

\noindent
We define discrete MDPs with rewards as the locale \texttt{MDP\_reward}.
\begin{IsabelleSnippet}{MDP Locale, label=snip:locale}
locale MDP_reward = discrete_MDP A K
  for
    A and 
    K :: "'s ::countable \<times> 'a ::countable \<Rightarrow> 's pmf" +
  fixes
    r :: "('s \<times> 'a) \<Rightarrow> real" and l :: real
  assumes
    zero_le_disc [simp]: "0 \<le> l" and
    disc_lt_one [simp]: "l < 1" and
    r_bounded: "bounded (range r)"
\end{IsabelleSnippet}

\noindent
The definition and $\nu_{\texttt{fin}}$ corresponds to $\nu_N$ in the paper.
\begin{IsabelleSnippet}{Expected Total Discounted Reward, label=snip:etr}
abbreviation "\<nu>_trace_fin t N \<equiv> \<Sum>i < N. l ^ i * r (t !! i)"
definition "\<nu>_fin n s = \<integral>t. \<nu>_trace_fin t n \<partial>\<T> p s"MDP
definition "\<nu> s = lim (\<lambda>n. \<nu>_fin n s)"
lemma \<nu>_eq_Pn: 
  "\<nu> s = (\<Sum>i. l^i * measure_pmf.expectation (Pn' p s i) r)"
\end{IsabelleSnippet}

\noindent
In our exposition, we use $\etropt$ for \texttt{$\nu$\_opt} and \texttt{$\nu_b$\_opt} .
\begin{IsabelleSnippet}{Optimal Reward, label=snip:etropt}
definition "\<nu>_opt s \<equiv> \<Squnion>p \<in> \<Pi>\<^sub>H\<^sub>R. \<nu> p s"
lemma \<nu>\<^sub>b_opt_eq_MR: 
  "\<nu>\<^sub>b_opt s = (\<Squnion>p \<in> \<Pi>\<^sub>M\<^sub>R. \<nu>\<^sub>b (mk_markovian p) s)"
\end{IsabelleSnippet}

\noindent
The constant $\PP$ in the paper is called $\mathcal{P}_\texttt{X}$ in the formalization.
\begin{IsabelleSnippet}{Vector Notation, label=snip:vector}
abbreviation "r_dec d s \<equiv> \<integral>a. r (s, a) \<partial>d s"
definition "\<P>\<^sub>X p n = push_exp (\<lambda>s. Xn' (mk_markovian p) s n)"
lemma \<nu>_eq_\<P>\<^sub>X: 
  "\<nu> (mk_markovian p) = (\<Sum>i. l^i *\<^sub>R \<P>\<^sub>X p i (r_dec\<^sub>b (p i)))"
\end{IsabelleSnippet}

\begin{IsabelleSnippet}{Value of Decision Rules,label=snip:inv}
lemma \<nu>_stationary: "\<nu>\<^sub>b (mk_stationary d) = 
  (\<Sum>t. l^t *\<^sub>R (\<P>\<^sub>1 d ^^ t)) (r_dec\<^sub>b d)"
  
lemma \<nu>_stationary_inv: "\<nu>\<^sub>b (mk_stationary d) = 
  inv\<^sub>L (id_blinfun - l *\<^sub>R \<P>\<^sub>1 d) (r_dec\<^sub>b d)"
\end{IsabelleSnippet}

\noindent
The Bellman operator $\cL_d$ is formally expressed as \texttt{L\,d}.
\begin{IsabelleSnippet}{Bellman operator,label=snip:bellman}
definition "L d v \<equiv> r_dec\<^sub>b d + l *\<^sub>R \<P>\<^sub>1 d v"
lemma \<nu>_step: "\<nu>\<^sub>b (mk_markovian p) = 
  L (p 0) (\<nu>\<^sub>b (mk_markovian (\<lambda>n. p (Suc n))))"
lemma L_\<nu>_fix_iff: "L d v = v \<longleftrightarrow> v = \<nu>\<^sub>b (mk_stationary d)"
\end{IsabelleSnippet}

\begin{IsabelleSnippet}{Bellman Optimality Operator, label=snip:bellmanopt}
definition "\<L> (v :: 's \<Rightarrow>\<^sub>b real) s = (\<Squnion>d \<in> D\<^sub>R. L d v s)"
lemma \<L>\<^sub>b_mono[intro]: "u \<le> v \<Longrightarrow> \<L>\<^sub>b u \<le> \<L>\<^sub>b v"

lemma step_mono_elem:
  assumes "v \<le> \<L>\<^sub>b v" "e > 0"
  shows "\<exists>d\<in>D\<^sub>R. v \<le> L d v + e *\<^sub>R 1"
\end{IsabelleSnippet}

\noindent
This proof shows that $\etropt$ is the unique fixed point of $\LL$.
\begin{IsabelleSnippet}{Solutions to the Optimality Equations, label=snip:sols}
lemma \<L>_dec_ge_opt:
  assumes "\<L>\<^sub>b v \<le> v"
  shows "\<nu>\<^sub>b_opt \<le> v"
proof -
  have "\<nu>\<^sub>b (mk_markovian p) \<le> v" if "p \<in> \<Pi>\<^sub>M\<^sub>R" for p
  proof -
    let ?p = "mk_markovian p"
    have aux: "\<nu>\<^sub>b_fin ?p n + l^n *\<^sub>R \<P>\<^sub>X p n v \<le> v" for n
      ...
    have 1: 
      "(\<lambda>n. (\<nu>\<^sub>b_fin ?p n + \<P>\<^sub>d p n v) s) \<longlonglongrightarrow> \<nu>\<^sub>b ?p s" for s 
        ...
    have "\<nu>\<^sub>b ?p s \<le> v s" for s ...
    thus ?thesis ...
  qed
  thus ?thesis ...
qed

lemma \<L>_inc_le_opt:
  assumes "v \<le> \<L>\<^sub>b v"
  shows "v \<le> \<nu>\<^sub>b_opt"
proof -
  have aux: "v s \<le> \<nu>\<^sub>b_opt s + (e/(1-l))" if "e > 0" for s e
  proof -
    obtain d where "d \<in> D\<^sub>R" and hd: "v \<le> L d v + e *\<^sub>R 1" ...
    let ?Pinf = "(\<Sum>i. l^i *\<^sub>R \<P>\<^sub>1 d^^i)"
    have "v \<le> r_dec\<^sub>b d + l *\<^sub>R (\<P>\<^sub>1 d) v + e *\<^sub>R 1"
    hence "(id_blinfun - l *\<^sub>R \<P>\<^sub>1 d) v \<le> r_dec\<^sub>b d + e *\<^sub>R 1"
    hence "?Pinf ((id_blinfun - l *\<^sub>R \<P>\<^sub>1 d) v) 
      \<le> ?Pinf (r_dec\<^sub>b d + e *\<^sub>R 1)" ...
    hence "v \<le> ?Pinf (r_dec\<^sub>b d + e *\<^sub>R 1)" ...
    also have 
      "\<dots> = \<nu>\<^sub>b (mk_stationary d) + e *\<^sub>R ?Pinf 1" ...
    finally have 
      "v s \<le> (\<nu>\<^sub>b (mk_stationary d) + (e/(1-l)) *\<^sub>R  1) s" ...
    thus "v s \<le> \<nu>\<^sub>b_opt s + (e/(1-l))" ....
  qed
  hence "v s \<le> \<nu>\<^sub>b_opt s + e" if "e > 0" for s e ...
  thus ?thesis ...
qed

lemma \<L>_fix_imp_opt:
  assumes "v = \<L>\<^sub>b v"
  shows "v = \<nu>\<^sub>b_opt"

lemma contraction_\<L>: "dist (\<L>\<^sub>b v) (\<L>\<^sub>b u) \<le> l * dist v u"
lemma \<L>\<^sub>b_fix_iff_opt [simp]: "\<L>\<^sub>b v = v \<longleftrightarrow> v = \<nu>\<^sub>b_opt"
lemma \<L>\<^sub>b_lim: "(\<lambda>n. (\<L>\<^sub>b ^^ n) v) \<longlonglongrightarrow> \<nu>\<^sub>b_opt"
\end{IsabelleSnippet}

\begin{IsabelleSnippet}{Existence of Optimal Policies, label=snip:ex}
lemma opt_imp_opt_dec_det:
  assumes "p \<in> \<Pi>\<^sub>H\<^sub>R" "\<nu>\<^sub>b p = \<nu>\<^sub>b_opt" 
  shows "\<exists>d \<in> D\<^sub>D. \<nu>\<^sub>b (mk_stationary_det d) = \<nu>\<^sub>b_opt"
\end{IsabelleSnippet}

\noindent
The definition \texttt{vi\_policy} corresponds to Algorithm \ref{alg:vi}.
\begin{IsabelleSnippet}{Value Iteration, label=snip:vi}
function 
  value_iteration :: "real \<Rightarrow> ('s \<Rightarrow>\<^sub>b real) \<Rightarrow> ('s \<Rightarrow>\<^sub>b real)" 
  where "value_iteration eps v = (
    if 2 * l * dist v (\<L>\<^sub>b v) < eps * (1-l) \<or> eps \<le> 0 
    then \<L>\<^sub>b v 
    else value_iteration eps (\<L>\<^sub>b v))"
definition 
  "find_policy (v :: 's \<Rightarrow>\<^sub>b real) s = arg_max_on (\<lambda>a. L\<^sub>a a v s) (A s)"
definition 
  "vi_policy eps v = find_policy (value_iteration eps v)"

lemma vi_policy_opt:
  assumes "0 < eps"
  shows 
    "dist (\<nu>\<^sub>b (mk_stationary_det (vi_policy eps v))) \<nu>\<^sub>b_opt 
      < eps"
\end{IsabelleSnippet}

\noindent
The definition \texttt{policy\_iteration} corresponds to Algorithm \ref{alg:pi}. 
\begin{IsabelleSnippet}{Policy Iteration, label=snip:pi}
definition "policy_eval d = \<nu>\<^sub>b (mk_stationary_det d)"
  
definition "policy_improvement d v s = (
  if is_arg_max 
    (\<lambda>a. L\<^sub>a a (apply_bfun v) s) (\<lambda>a. a \<in> A s) (d s) 
  then d s
  else arb_act (opt_acts v s))"
  
definition "policy_step d = 
  policy_improvement d (policy_eval d)"
  
function policy_iteration :: "('s \<Rightarrow> 'a) \<Rightarrow> ('s \<Rightarrow> 'a)" where
  "policy_iteration d = (
  let d' = policy_step d in
  if d = d' \<or> \<not>is_dec_det d then d else policy_iteration d')"

lemma policy_iteration_correct: 
  "d \<in> D\<^sub>D \<Longrightarrow> \<nu>\<^sub>b (mk_stationary_det (policy_iteration d)) = \<nu>\<^sub>b_opt" 
\end{IsabelleSnippet}

\noindent
We introduce a proper locale for splitting methods and show that this variant of value iteration converges.
\begin{IsabelleSnippet}{Splitting Methods, label=snip:splitting}
definition "is_splitting_blin X Q R \<longleftrightarrow>
  X = Q - R \<and> invertible\<^sub>L Q 
    \<and> nonneg_blinfun (inv\<^sub>L Q) \<and> nonneg_blinfun R"

abbreviation "L_split d v \<equiv> 
  inv\<^sub>L (Q d) (r_dec\<^sub>b (mk_dec_det d) + R d v)"
definition "\<L>_split v s = (\<Squnion>d \<in> D\<^sub>D. L_split d v s)"
\end{IsabelleSnippet}

\noindent
The definition \texttt{vi\_gs\_policy} corresponds to Algorithm \ref{alg:vigs}. 
\begin{IsabelleSnippet}{Gauss-Seidel Value Iteration, label=snip:gs}
lift_definition 
  \<P>\<^sub>U :: "(nat \<Rightarrow> nat) \<Rightarrow> (nat \<Rightarrow>\<^sub>b real) \<Rightarrow>\<^sub>L nat \<Rightarrow>\<^sub>b real" 
    is "\<lambda>d (v :: nat \<Rightarrow>\<^sub>b real). 
      (Bfun (\<lambda>s. (\<P>\<^sub>1 (mk_dec_det d) 
        (bfun_if (\<lambda>s'. s' < s) 0 v) s)))"

definition "Q_GS d = id_blinfun - l *\<^sub>R \<P>\<^sub>L d"
definition "R_GS d = l *\<^sub>R \<P>\<^sub>U d"
lemma splitting_gauss: "is_splitting_blin 
  (id_blinfun - l *\<^sub>R \<P>\<^sub>1 (mk_dec_det d)) (Q_GS d) (R_GS d)"

definition "GS_inv d v = 
  inv\<^sub>L (Q_GS d) (r_dec\<^sub>b (mk_dec_det d) + R_GS d v)"
lemma GS_inv_rec: "GS_inv d v = 
  r_det\<^sub>b d + l *\<^sub>R (\<P>\<^sub>U d v + \<P>\<^sub>L d (GS_inv d v))"
lemma GS_indep_high_states:
  assumes "\<And>s'. s' \<le> s \<Longrightarrow> d s' = d' s'"
  shows "GS_inv d v s = GS_inv d' v s"

lemma ex_GS_arg_max_all: 
  "\<exists>d. is_arg_max (\<lambda>d. GS_inv d v s) (\<lambda>d. d \<in> D\<^sub>D) d"
\end{IsabelleSnippet}

\noindent
The definition \texttt{mpi\_algo} corresponds to Algorithm \ref{alg:mpi}. 
\begin{IsabelleSnippet}{Modified Policy Iteration, label=snip:mpi}
definition "L_pow v d m = (L (mk_dec_det d) ^^ m) v"

fun mpi :: "nat \<Rightarrow> (('s \<Rightarrow> 'a) \<times> ('s \<Rightarrow>\<^sub>b real))" where
  "mpi 0 = (policy_improvement d0 v0, v0)" |
  "mpi (Suc n) =
  (let (d, v) = mpi n; v' = L_pow v d (Suc (m n v)) in
  (policy_improvement d v', v'))"

definition "mpi_val n = snd (mpi n)"

theorem mpi_conv:
  assumes "v0 \<le> \<L>\<^sub>b v0"
  shows "mpi_val \<longlonglongrightarrow> \<nu>\<^sub>b_opt" 
  and "\<And>n. mpi_val n \<le> mpi_val (Suc n)"
\end{IsabelleSnippet}

\section{Appendix: Running Time Data}
\label{sec:rtd}

\begin{figure}[h]
  \includegraphics[width=.9\linewidth]{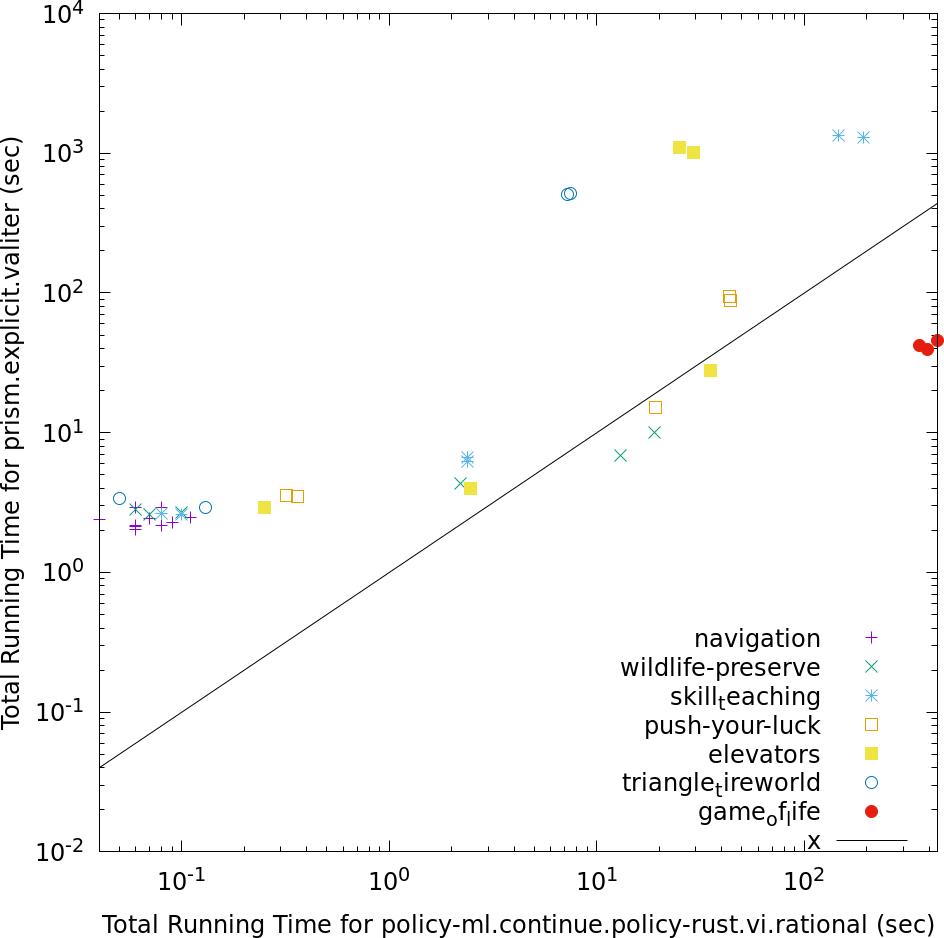}
  \caption{Running times of combined VI vs.\ PRISM value iteration.}
  \label{fig:sub1}
\end{figure}

\begin{figure}[h]
  \includegraphics[width=.9\linewidth]{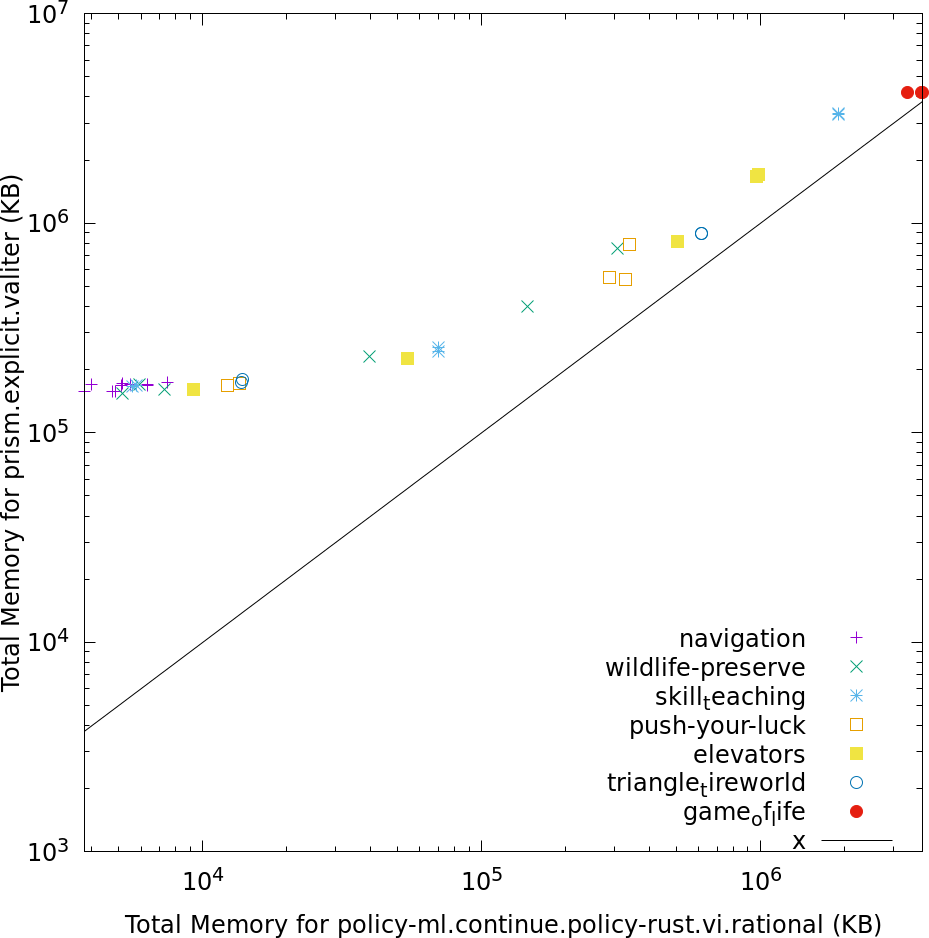}
  \caption{Memory usage of combined VI vs.\ PRISM value iteration.}
  \label{fig:sub2}
\end{figure}

\end{document}